\documentclass[11pt,a4paper]{article}
\usepackage[hyperref]{emnlp-ijcnlp-2019}
\usepackage{times}
\usepackage{latexsym}
\usepackage{booktabs}
\usepackage{graphicx}
\usepackage{caption}
\usepackage{url}
\usepackage{amsmath}
\usepackage{graphicx}
\usepackage{float}
\aclfinalcopy % Uncomment this line for the final submission

\usepackage{lipsum}

\newcommand\blfootnote[1]{%
  \begingroup
  \renewcommand\thefootnote{}\footnote{#1}%
  \addtocounter{footnote}{-1}%
  \endgroup
}
\title{Author2Vec: A Framework for Generating User Embedding}

\author{Xiaodong Wu* $\;\;\;\;$   Weizhe Lin*  $\;\;\;\;$ Zhilin Wang $\;\;\;\;$  Elena Rastorgueva\\
  University of Cambridge, United Kingdom \\
  {\tt \{xw338, wl356, zw322\}@cam.ac.uk,} \\
  {\tt elenaras@cantab.net}
  }
\date{}

\begin{document}
\maketitle
\begin{abstract}
\blfootnote{*Equally contribute to this work}
Online forums and social media platforms provide noisy but valuable data everyday. In this paper, we propose a novel end-to-end neural network based user embedding system, Author2Vec. The model incorporates sentence representations generated by BERT (Bidirectional Encoder Representations from Transformers) \cite{devlin2018bert} with a novel unsupervised pre-training objective, authorship classification, to produce better user embedding that encodes useful user-intrinsic properties. This user embedding system was pre-trained on post data of 10k Reddit users and was analyzed and evaluated on two user classification benchmarks: depression detection and personality classification, in which the model proved to outperform traditional count-based and prediction-based methods. We substantiate that Author2Vec successfully encoded useful user attributes and the generated user embedding performs well in downstream classification tasks without further finetuning.
\end{abstract}

\section{Introduction}
With the rising popularity of various social media, there is also a rising need for understanding social media users. In recent years, natural language processing (NLP) has gained increasing popularity and is now widely used in many natural language understanding tasks. Capable language models, such as BERT and XLNet \cite{yang2019xlnet}, have been developed recently. These new technologies can enable the analysis of additional features of social media users from the user-generated textual data.

Much work have shown that NLP technologies can be used to understand the demography of social media \cite{xu2012modeling}, political leaning \cite{kosinski2013private, pennacchiotti2011machine, schwartz2013personality}, emotions \cite{ofoghi2016towards}, personality and sexual orientation \cite{kosinski2013private} of the users.

In terms of mental status of social media users, \newcite{mitchell2015quantifying} showed that linguistic traits are predictive of schizophrenia, while \newcite{preoctiuc2015role} investigated the link between personality types, social media behavior, and psychological disorders, such as depression and Post Traumatic Stress Disorder (PTSD). They suggest that certain personality traits are correlated to mental illnesses.

User profiling has become one of the hottest research topics in social media
analysis, and has been applied to various domains, such as discovering potential business customers and generating intelligent marketing reports for different brands \cite{li2019improving}. Latent user characteristics such as brand preferences \cite{pennacchiotti2011machine} are also of great importance for marketing.

Given the intererest in understanding social media users, we investigated how recent NLP developments could be used to facilitate social media user analysis. We aimed at creating an end-to-end user embedding system to generate effective and discriminative embedding for social media users themselves. We expected that even without further finetuning or feature engineering on user post corpora for specific classification tasks, our pre-trained framework could still showcase a good performance in classification for unseen users and their posts.

The main contributions of this paper are:\\
1. We proposes an end-to-end framework for user embedding generation, which was built upon the sentences embedding generated by a BERT model.\\
2. We evaluate our generated user embedding on several existing social media user benchmarks and compared the classification ability of our user embedding with that of other baseline models.
\section{Related Work}
\subsection{Social media user modelling and attribute classification}
Much work has been done to model the behaviors of social media users. Most of the existing user modelling methods build profiles for each user based on their tweets or posts by extracting key words \cite{chen2010short}, entities \cite{abel2011analyzing}, categories \cite{michelson2010discovering} or latent topics \cite{hong2010empirical}. \newcite{xu2012modeling} incorporated three factors into modelling  social media users: breaking news happening at that moment, posts published by their friends recently and their intrinsic interests. However, their examination focused more on the posting behaviors of social media users rather than general author attributes.
 
In terms of attribute classification, previous research has mostly focused on feature engineering \cite{mueller2016gender, alowibdi2013language, sloan2015tweets}. However, feature engineering generally requires a lot of manual labor to design and extract task-specific features. Moreover, to achieve the ideal performance, different features are extracted for different attributes, which may limit the scalability of learned classification models \cite{li2019improving}. Many of these studies were based on traditional machine learning classifiers \cite{rahimi2015exploiting,  sesa2016gender, volkova2015inferring}. However, recently some neural network methods were also adopted to construct a large framework combining different features. For example, \newcite{li2019improving} proposed a complex neural network with an attention mechanism to incorporate a text-based embedding and a network embedding characterising the social engagement of the users.
 
 \subsection{Text-based user embedding}
The purpose of a text-based user embedding is to map a sequence of social media posts by the same author into a vector representation which captures the linguistic or higher-level features expressed in the text. There are many practical methods for encoding the users, which will be explored in the following sections.

\subsubsection{Count-based methods}
Latent Dirichlet Allocation (LDA) \cite{blei2003latent} is a popular generative graphical model for embedding generation \cite{schwartz2013personality, hu2016language}. Latent Semantic Analysis / Indexing (LSA/LSI) \cite{deerwester1990indexing} using Singular Value Decomposition (SVD) and Principle Component Analysis (PCA) has also been used \cite{kosinski2013private}. These are Bag-of-Words models which characterise the topic composition of the posts.

\subsubsection{Prediction-based methods}
Word2Vec \cite{mikolov2013efficient}, as well as Global Vectors for Word Representation (GloVe) \cite{pennington2014glove} and other variants, are popular neural network-based models for extracting dense vector representation of words. They have been used in many NLP applications including generating user embedding, where all the word vectors are aggregated by methods such as averaging \cite{benton2016learning, ding2017multi}. Its extension, Doc2Vec \cite{le2014distributed}, can generate dense low dimensional vectors for a document while the Paragraph Vector Model \cite{le2014distributed} is an alternative choice for researchers \cite{song2017learning, yu2016user}. There are two typical methods for learning a user embedding from such vector representations: either concatenating all the posts from the same user (User-D2V), or simply deriving a user embedding from all the post vectors from the same person using some pooling methods (Post-D2V).

Furthermore, Recurrent Neural Network (RNN) models such as Long Short-Term Memory (LSTM) \cite{hochreiter1997long} were also used to capture temporal information, for example, the posting order of user posts \cite{zhang2018user}.

\subsection{Social media depression detection}
\newcite{pirina2018identifying} proposed a promising way to prepare and collect data for a social media user depression classification task. They evaluated the classification task using several different configurations of linear Support Vector Machines (SVMs) with bag-of-n-grams (BON) \cite{bespalov2011sentiment} features.

\subsection{Social media personality classification}

The Myers Briggs Type Indicator (MBTI) \cite{myers1998mbti} is a personality type system that groups personalities into 16 distinct types across 4 axes, shown in Table \ref{mbti_table}.

\begin{table}[http!]
\centering
\begin{tabular}{l|r}
    \toprule
    Axis Group 1 & Axis Group 2\\
    \hline
    Introversion (I) & Extroversion (E)\\
    Intuition (N) & Sensing (S)\\
    Thinking (T) & Feeling (F)\\
    Judging (J) & Perceiving (P)\\
    \bottomrule
\end{tabular}
\caption{MBTI Types}
\label{mbti_table}
\end{table}

\newcite{gjurkovic2018reddit} introduced a new, large-scale dataset MBTI9k. They carefully scrapped around 9,700 Reddit users and labeled their MBTI personality types. To classify the users' personality, they utilised and engineered many different features such as user activity and posting behavior features, type-token ratio, and LIWC features \cite{pennebaker2015development}.

\section{Author2Vec User Embedding System\label{model_section}}
\begin{figure*}[htp]
    \centering
    \includegraphics[width=16cm]{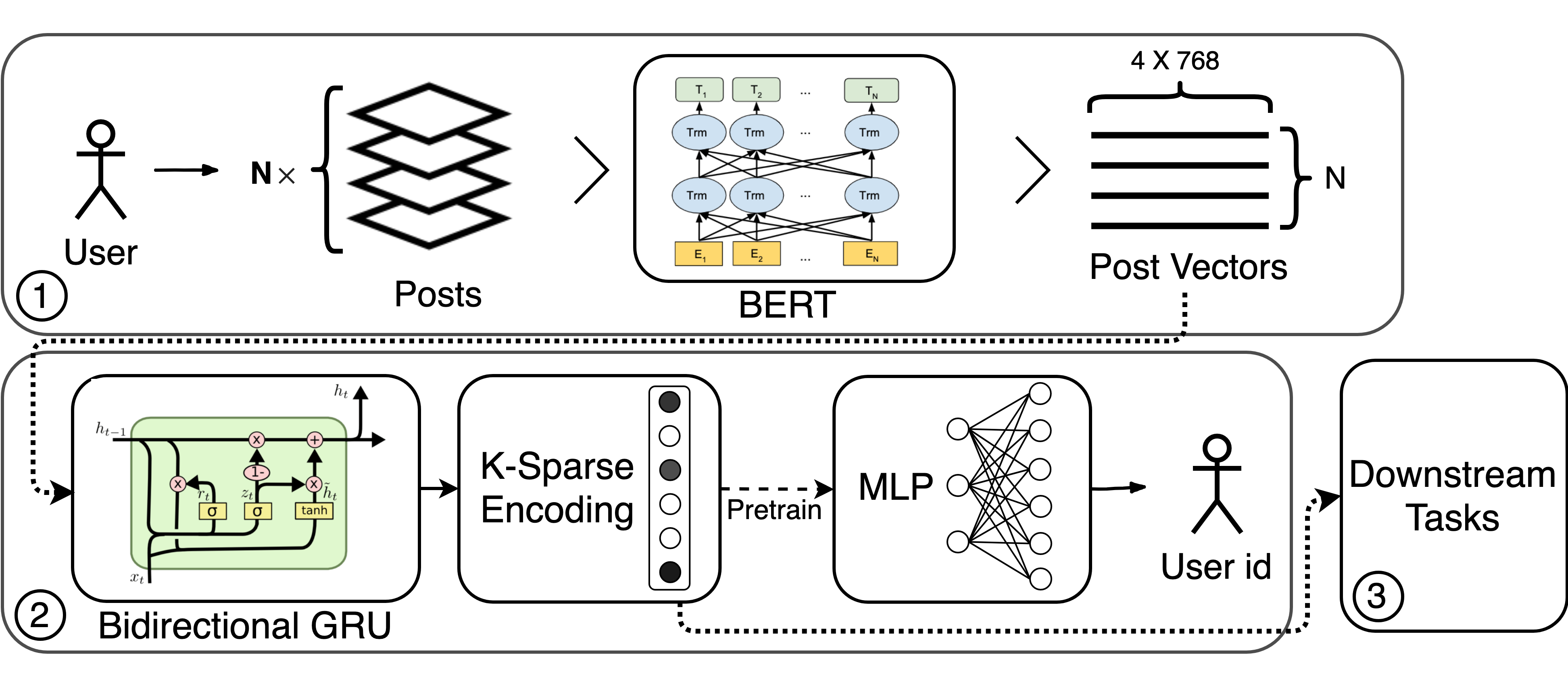}
    \caption{Three Stages of Proposed Author2Vec System:    1) Convert user' posts to post embedding. 2) pre-train Author2Vec on authorship classification.    3)  Apply user embedding to downstream tasks}
    \label{fig:model_structure}
\end{figure*}
In this section, we described the pipeline we followed to build the Author2Vec model for Reddit users, as was shown in Figure \ref{fig:model_structure}.
\subsection{Data collection and preprocessing\label{preprocessing}}
We scraped 340,000 active users' posts (users who had posted more than 20 posts) from several subreddits \footnote{e.g. r/depression, r/relationship\_advice, r/offmychest, r/IAmA, r/needadvice, r/tifu, r/confessions, r/confession, r/TrueOffMyChest, r/confidence, r/socialanxiety, r/Anxiety, r/socialskills, r/happy} using the Pushshift Reddit API (\url{https://github.com/pushshift/api}). We randomly picked 10,000 as our experimental users in this section. For each selected user, we would scrape their most recent 500 posts as the input to our system.

A pre-trained BERT model (the Base Uncased 12-layer, 768-hidden, 12-heads, 110M parameters model) was used to extract post representations.

Each post was tokenized using the BERT Byte-Pair Encoding tokenizer, and was ruled out if it:

1. contained a large portion of non-textual or meaningless data, for example repetitive letters or a single picture/video link, or

2. contained fewer than 20 tokens.

\begin{figure}[h]
    \centering
    \includegraphics[width=8cm]{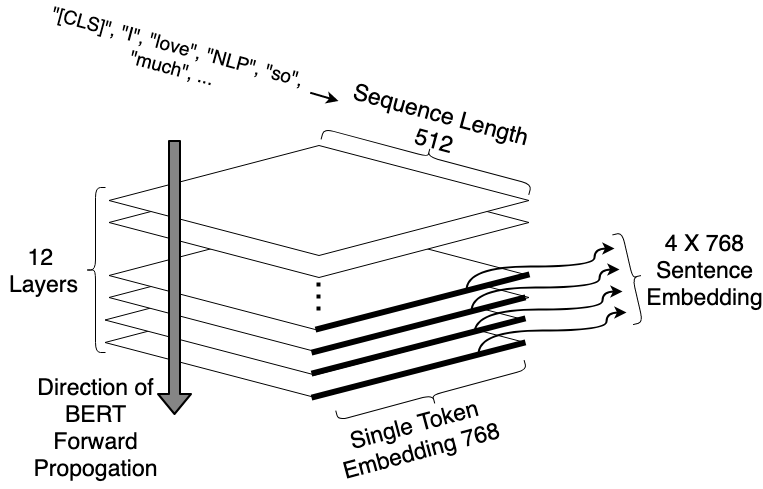}
    \caption{Embedding Extraction from BERT}
    \label{fig:bert_extraction}
\end{figure}

The posts were then fed into the BERT pipeline powered by bert-as-service \cite{xiao2018bertservice}. For each post, 12 layers of 512 (time dimension) $\times$ 768 (feature dimension) embedding were generated. We concatenated the first token ([CLS]) representation of the last four layers as the post representation as was shown in Figure \ref{fig:bert_extraction}, which we expected to give a good encoding of the semantics of a post based on the findings in \cite{devlin2018bert}.

After pre-processing, we gained a list of 3072-dimension embedding vectors to represent each author, which would be fed into the Author2Vec system as input.

\subsection{Authorship classification pre-training}
To obtain an text-based end-to-end embedding model that can extract a good encoding of an author based on their posts, we propose a novel unsupervised pre-training objective, authorship prediction. 

During pre-training, a MLP classifier would be attached to the embedding model to allow classification. Each training sample contained a subset of the posts of a randomly-picked author. The model was trained to predict the author of the posts. 

After pre-training stage, the MLP classifier could be removed to obtain the target embedding model.

\subsection{Model Architecture}
Overall, our embedding model comprises of a 512 units Bidirectional Gated Recurrent Unit (GRU) \cite{cho2014properties} that converted a variable number of post embedding of the same author to a fixed-length vector, followed by a 768 units linearly activated K-Sparse encoding layer \cite{makhzani2013k} that could learn to give sparse encoding of the an author. During the pre-training stage, a MLP of 256 units ReLU-activated hidden layers would be attached to the K-Sparse encoding layer to classify a number of different authors. During the inference stage, the K-Sparse outputs would be used directly to give user embedding.
The GRU mentioned here is a gating mechanism in a recurrent neural network, which allows the use of fewer parameters and leads to faster convergence \cite{chung2014empirical}.
% It is formulated as follows:
% \begin{align*}
% z_{t} & =\sigma\big(x_{t}U^{z}+h_{t-1}W^{z}\big)\\
% r_{t} & =\sigma\big(x_{t}U^{r}+h_{t-1}W^{r}\big)\\
% \tilde{h}_{t} & =\tanh\big(x_{t}U^{h}+(r_{t}\ast h_{t-1})W^{h}\big)\\
% h_{t} & =(1-z_{t})\ast h_{t-1}+z_{t}\ast\tilde{h}_{t}
% \end{align*}
% Where $r$ is a reset gate, and $z$ is an update gate. The reset gate determines how to combine the new input with the previous memory, and the update gate defines how much of the previous memory to maintain. 
Making GRU Bidirectional wrapper allows information to flow along both directions, which should further boost performance.

The K-sparse encoding layer aforementioned is a layer which allows only the $k$ most significant values to pass while the other insignificant values will be set to zero. This mechanism could yield more semantic features and act as a good regularization against overfitting. In our model, the sparsity level $k$ was set to 32 during pre-training stage and 64 during inference stage.

\subsection{Baseline LSI Author2Vec}
We used the traditional count-based method Latent Semantic Indexing (LSI) as our baseline post embedding model for comparison.

Gensim (\url{https://radimrehurek.com/gensim}) was used to implement LSI. After removing tokens that were contained in fewer than 10 posts or in more than 30\% of all posts, term frequency-inverse document frequency (TF-IDF) \cite{salton1988term} was applied on each post before LSI to boost performance. Finally a vector of length 500 was calculated for each post. We then used the same pipeline as BERT-based Author2Vec pre-training to build an LSI-based Author2Vec model for comparison.

To make the comparison fair, we used the same preprocessing method for both the LSI-based system and the BERT-based system.

\subsection{Evaluation}
To evaluate and compare the performance in ``authorship'' classification, we trained a smaller version of the Author2Vec model for both LSI post embedding and BERT post embedding on a subset of users. We picked 3000 Reddit users who had more than 80 valid posts. For each user, 40 of their posts were used to generate test data, and the rest were used to generate training data. We fixed the training and testing partitions which means the training posts and the testing posts were the same in each system. \textbf{Note that the test data here was purely for evaluation purpose. During actual pre-training of Author2Vec, to fully utilize the available data, all posts would be used to train the model and test data would be absent.}

As was shown in Table \ref{table_pretrain}, even though the LSI model obtained a higher performance on the training set, it overfitted significantly on the test set. On the other hand, the BERT representation achieved 10\% higher top-5 and top-1 accuracy on the test set, which demonstrated that the BERT based Author2Vec representation could encode more distinguishing features of a large number of users and is more suitable for our purpose.

\begin{table}[h]
\begin{tabular}{lccc}
\toprule
Model & Partition & Accuracy & Top-5 acc.\\
\hline
LSI\_500 & training & \textbf{95.73} & \textbf{99.89}\\
BERT\_3072 & training & 95.14 & 99.43\\
\hline
LSI\_500 & test & 65.43 & 81.47\\
BERT\_3072 & test & \textbf{74.22} & \textbf{91.21}\\
\bottomrule
\end{tabular}\\
\caption{Accuracy and top-5 accuracy of training and test set for different models (best performance in \textbf{bold})}
\label{table_pretrain}
\end{table}

\section{Preliminary Embedding Evaluation}
To gain a preliminary understanding of user embedding generated by Author2Vec, we visualized and then quantitatively evaluated the 768-dimension sparse embedding via a simple task: gender classification.

\textbf{In this and all the following sections}, the Author2Vec model was pre-trained on 10,522 different Reddit users, using the proposed preprocessing methods and the embedding system described in Section \ref{model_section}.

\subsection{Dataset}
``Gender statistics of /r/RateMe'' (\url{https://www.kaggle.com/nikkou/gender-statistics-of-rrateme}) was a database that collected and parsed the posts on ``Rateme'' subreddit where people posted their age, gender together with their selfies, welcoming ratings from other social media users. The labels included but are not limited to gender and age parsed from the posts. The dataset contained around 295,000 posts and there were 4,991 active authors who had more than 20 posts in their accounts. Their recent posts (up to 500 posts) were collected and pre-processed as we mentioned in Section \ref{preprocessing}. 
After removing the authors with unknown gender (failed to parse), 4802 authors remained in our database, among which there were 4073 males and 729 females. Note that these authors do not overlap with the users in pre-training stage.
 
\subsection{Visualisation}

\begin{figure}[htp]
    \centering
    \includegraphics[width=6cm]{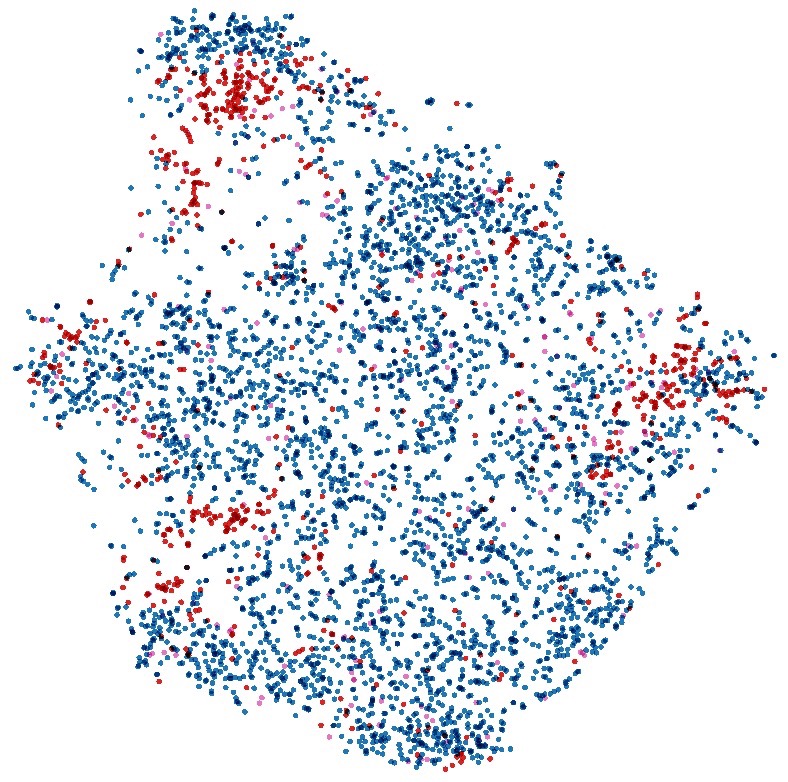}
    \caption{Visualisation of user embedding labelled with gender (blue for male and red for female)}
    \label{fig:gender_visualisation}
\end{figure}

Using the pre-trained Author2Vec model, the embedding sequence of the posts of each author (dimension: number of posts $\times$ 3072) were transformed into a user embedding vector (dimension: 1 $\times$ 786).

t-Stochastic Neighbor Embedding (t-SNE) \cite{maaten2008visualizing} is a convenient tool for automated dimension reduction and visualisation. It transfers the data similarity into probability and then projects the data onto a lower dimension space. We plotted the user embedding onto a 2-D graph using t-SNE, demonstrating the ability of our embedding to distinguish between different genders. As was shown in Figure \ref{fig:gender_visualisation}, the red points (females) formed three clear clusters. This gave the intuition that the user embedding generated by our pre-trained Author2Vec successfully encoded some intrinsic properties of unseen users, in this case, gender information.

\subsection{Validation\label{gender_visualisation_section}}
In order to validate the intuition given by the visualisation step and evaluate the ability of generalisation on gender classification, we fed the user embedding of each user into an MLP with one ReLU activated 256-dense hidden layer to predict gender of unseen Reddit users. A 10-fold cross validation was performed during evaluation. To evaluate the performance of our model under extreme conditions, we also tried \textbf{reversed 10-fold: train our model with only one fold and test it on all the remaining nine folds of data}. To correctly reflect the classification performance on an unbalanced dataset (male-female ratio of 5.59), the average value and standard deviation of the weighted F1-score of each cross validation were reported.

The results of 10-fold and 10-fold-reverse cross validation were shown in Table \ref{table_gender}. In reverse 10-fold, even with very little training data (480 users), the MLP classification result based on Author2Vec user embedding gave a weighted F1-score of as high as 0.897. This result demonstrated that the pre-trained embedding model could give very robust and discriminative author embedding even on unseen users.

\begin{table}[h]
\begin{tabular}{lcccc}
\toprule
& \multicolumn{4}{c}{F1-score}\\
\cline{2-5}
& Min. & Max. & Avg. & Std.\\
\hline
10-fold & 0.907 & 0.948 & 0.933 & 0.010\\
10-fold reverse & 0.887 & 0.910 & 0.897 & 0.007\\
\bottomrule
\end{tabular}
\caption{10-fold and 10-fold reverse cross validation results of gender classification}
\label{table_gender}
\end{table}

\section{Benchmark Evaluation}
To further explore the potential of our user embedding system, we evaluated our system on two Reddit-based user classification benchmarks: Depression Detection and MBTI Personality Classification. For each benchmark, we built several baseline models to compare with our Author2Vec model.

\subsection{Baseline embedding model\label{pre-trained-models}}
Overall, we designed three baseline methods to generate user embedding: LSI, LDA and Word2Vec methods. 

1. LSI and LDA: For each user, we concatenated all their posts into one large document, and applied LSI or LDA on this document to generate a proxy embedding for the user. This embedding was then used for downstream classification. We implemented embedding dimensions of both 300 and 500 for both methods.

2. Word2Vec: We took the average of all word vectors of all the words in all the posts of each user and use this vector as a proxy embedding for the user. We used the Facebook FastText (\url{https://fasttext.cc/}) pre-trained Word2Vec model: crawl-300d-2M, which is a model with 2 million word vectors trained on Common Crawl (600B tokens).

The baseline implementations were different for each benchmark and are introduced in more detail in their respective benchmark sections.

\subsection{Depression detection}
In this benchmark, we tried to predict whether a user was depressed or not based on their recent posts.
\subsubsection{Dataset}
\newcite{pirina2018identifying} suggested that careful selection of the depression data source was important for not obtaining illusionary results for depression classification tasks. Therefore, the following steps were conducted in order to obtain an accurate dataset:

1. We scraped 4,500 authors who had posted in the ``r/depression'' subreddit, and then collected all of their posts.

2. We manually labelled 3,000 depressed authors according to their posts under the ``r/depression'' subreddit. Two researchers filtered the data and the inter-rater reliability was ensured by dropping all the authors without achieved consensus.

3. In order to prevent models from learning to attend to depression-related keywords instead of learning the semantics or style of users' posts to detect depression, we removed all the posts that directly mentioned depression related expressions such as "depression", "depressed" and "anxiety" and all the posts under depression related subreddits such as "r/Depression", "r/AskDoc" and "r/mentalhealth". 

4. We collected 3,000 non-depressed authors from the most popular non-depression-related subreddits \footnote{e.g. ``r/funny'', ``r/gaming'', ``r/science'' and ``r/AskReddit''} who did not post any depression-related posts. 

Finally, we obtained a dataset of 3000 depressed authors and 3000 non-depressed authors \textbf{without any overlap with pre-training data}. The labels were generally convincing given our careful selection process.
\subsubsection{Baseline\label{BOW_baseline}}
We implemented all three baseline methods described in Section \ref{pre-trained-models} ``LSI", ``LDA", ``Word2Vec''). For LSI and LDA models, we implemented versions that were only trained on the 10k user dataset in Section \ref{model_section} and versions that were trained directly on this 6k user depression dataset. Those models pre-trained on the 10k dataset were denoted with a prefix "Pre-trained-".
\subsubsection{Visualisation}
\begin{figure}[htp]
    \centering
    \includegraphics[width=5cm]{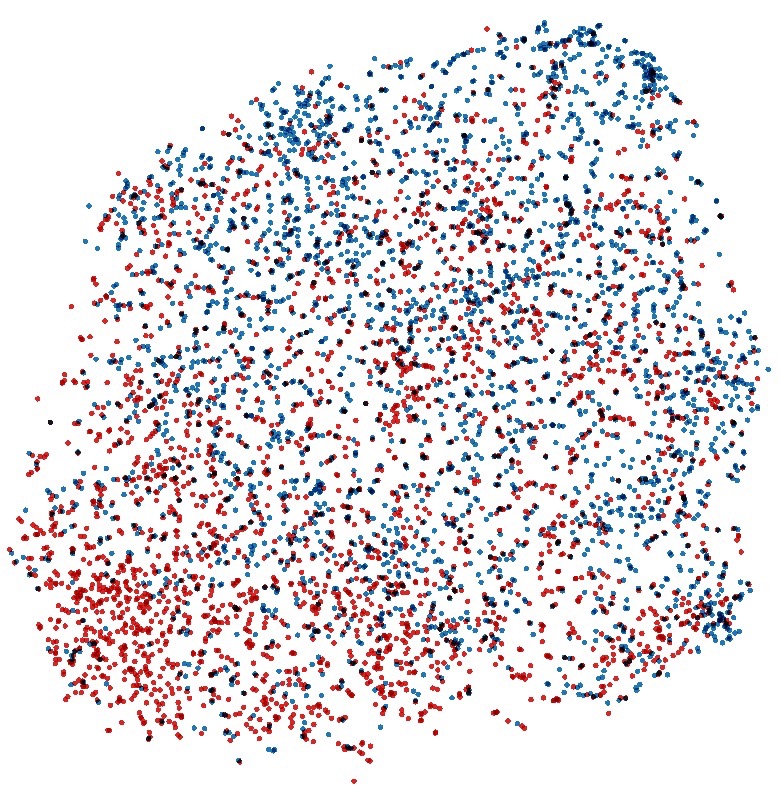}
    \caption{Visualisation of user embedding labelled with depression (blue for non-depression and red for depression)}
    \label{fig:depession_visualisation}
\end{figure}
We used the visualization method mentioned in Section \ref{gender_visualisation_section} to visualize the embedding generated for depression dataset users. Figure \ref{fig:depession_visualisation} showed a clear polarization of depressed and non-depressed author embedding, which implied that pre-trained Author2Vec successfully captured depression-related intrinsic user attributes.
\subsubsection{Evaluation}
\begin{table}[t]
\begin{center}
    5-fold Cross Validation F1-score\\
\end{center}

\begin{tabular}{lcc}
\toprule
Model & Avg. & Std.\\
\hline
LR Pre-trained-TF-IDF-LSI\_300 & 0.682 & 0.012\\
LR Pre-trained-TF-IDF-LSI\_500 & 0.679 & 0.009\\
LR TF-IDF-LSI\_300 & 0.683 & 0.009\\
LR TF-IDF-LSI\_500 & 0.681 & 0.009\\
LR Pre-trained-LDA\_300 & 0.659 & 0.011\\
LR Pre-trained-LDA\_500 & 0.667 & 0.015\\
LR LDA\_300 & 0.661 & \textbf{0.008}\\
LR LDA\_500 & 0.669 & 0.015\\
LR Word2Vec\_300 & 0.653 & 0.010\\
\midrule
Proposed Model\\
\hline
LR Author2Vec\_768 & 0.702 & 0.015\\
MLP Author2Vec\_768 & \textbf{0.720} & 0.015\\
\bottomrule
\end{tabular}
\caption{Comparison of the baseline and proposed user embedding for depression classification task (best results in \textbf{bold}). [LR denotes logistic regression, MLP denotes multilayer perceptron.]}
\label{depression_table}
\end{table}
We performed 5-fold cross validation on both Author2Vec and other baseline user embedding with a logistic regression model. In addition, an MLP with two ReLU activated hidden layers was used to further improve the performance of Author2Vec. The results were shown in Table \ref{depression_table}.

Among all the baseline models, LSI\_300 gave the highest F1-score of 68\%. However, the proposed Author2Vec embedding outperformed all the baseline embedding by at least 2\% and the performance was further improved to 72\% by a tuned MLP network. The result suggested that Author2Vec could encode other useful information that cannot be fully captured by count-based methods.

\subsection{Personality type classification}
\subsubsection{Dataset}
\newcite{gjurkovic2018reddit} introduced MBTI9k, a Reddit author dataset with convincing MBTI personality type labels (e.g. ``INTP'', explained in Table \ref{mbti_table}). They carefully decided the labels based on users' flair history (a small banner associated with its author). They also put in manual efforts to collect more authors of less popular MBTI types (ESFJ and ESTJ). The authors with non-unique MBTI type flairs were also ruled out. To avoid models making personality classifications by memorizing keywords, they removed all comments under 122 subreddits that revolved around MBTI-related topics and also those with MBTI-related content.

Due to their careful data selection and label filtering, we chose to evaluate our user embedding on this MBTI9k dataset. However, to keep the consistency of our experiment and maximise the potential of our proposed model which was pre-trained on author posts only, we used only author posts instead of using both posts and comments. We also ruled out less active authors who had fewer than 10 posts under their accounts. The filtered MBTI9k type distribution is shown in Table \ref{table_mbti}.

\begin{table}[]
    \centering
    \begin{tabular}{c|c|c|c}
    \toprule
    type & number & type & number\\
    \hline
    ISFJ & 6 & ESFJ & 6\\
    ISTJ & 14 & ESTJ & 8\\
    ISFP & 19 & ESFP & 11\\
    ISTP & 197 & ESTP & 101\\
    INFJ & 34 & ENFJ & 14\\
    INTJ & 97 & ENTJ & 41\\
    INFP & 214 & ENFP & 175\\
    INTP & 2801 & ENTP & 1361\\
    \bottomrule
    \end{tabular}
    % \begin{tabular}{c|c|c|c}
    % \toprule
    % E/I & S/N & T/F & J/P\\
    % \hline
    % 3382/1717 & 362/4737 & 4620/479 & 220/4879 
    % \bottomrule
    % \end{tabular}
    \caption{MBTI type distribution after filtering}
    \label{table_mbti}
\end{table}

\subsubsection{Baseline}
In this benchmark, we used ``Pre-trained-TF-IDF-LSI'', ``Pre-trained-LDA'' and ``Word2Vec'' as our baseline embedding model.

\subsubsection{Evaluation}
\begin{figure*}[h] 
  \begin{minipage}[b]{0.5\linewidth} 
    \centering
    \begin{tabular}{lcccc}
    \toprule
    & \multicolumn{4}{c}{F1 Score on Dimensions} \\
    \cline{2-5}
    Model & E/I & S/N & T/F & J/P\\
    \hline
    LR Word2Vec\_300 & 0.548 & 0.593 & 0.610 & 0.504\\
    LR TF-IDF-LSI\_300 & 0.613 & 0.692 & 0.672 & 0.607\\
    LR LDA\_300 & 0.566 & 0.651 & 0.658 & 0.554\\
    LR TF-IDF-LSI\_500 & 0.611 & 0.698 & 0.676 & 0.606\\
    LR LDA\_500 & 0.606 & 0.639 & 0.648 & 0.574\\
    \midrule
    Proposed Model\\
    \hline
    LR Author2Vec\_768 & \textbf{0.690} & \textbf{0.766} & \textbf{0.681} & \textbf{0.610}\\
    \bottomrule
    \end{tabular}\\
    \captionof{table}{Comparison of the baseline and proposed user embedding for MBTI type classification task (best results in \textbf{bold}) [LR denotes logistic regression.]}
    \label{table_personality}
  \end{minipage} 
  \begin{minipage}[b]{0.6\linewidth} 
    \centering 
    \includegraphics[width=7cm]{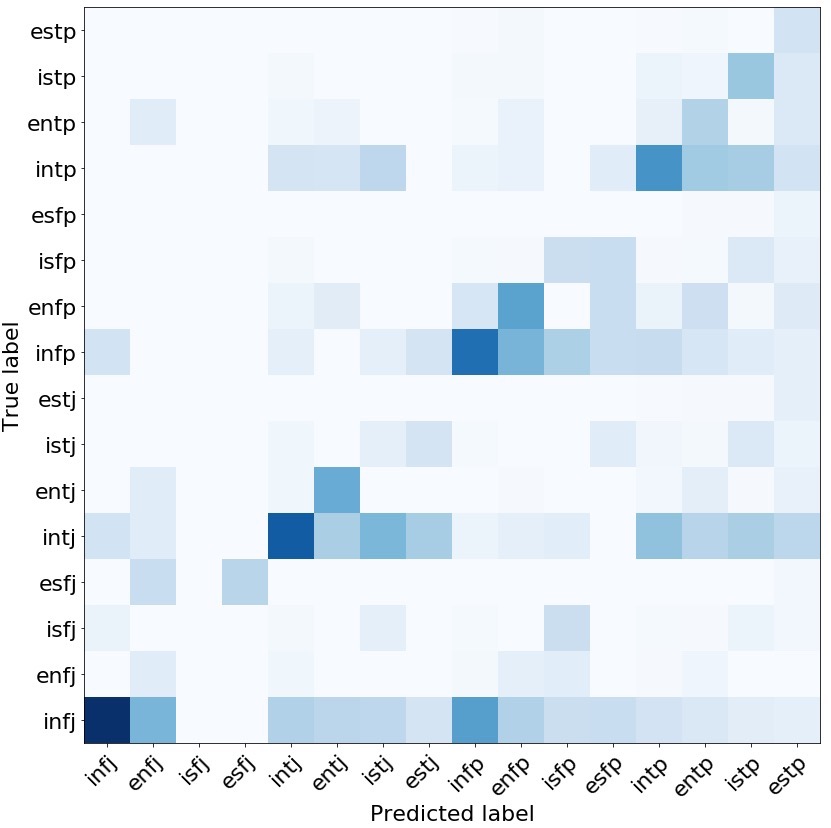}
    \caption{Heatmap of the confusion matrix} 
    \label{fig_heatmap} 
  \end{minipage}% 
\end{figure*}
Because some of the less popular MBTI types (e.g. ESTJ and ISFJ) had only a small number of authors, we chose to perform binary classification on each axis of MBTI types. We reported F1-score instead of accuracy in order to better characterise the model performance on unbalanced dataset.

As was shown in Table \ref{table_personality}, the proposed Author2Vec model outperformed all the other baseline models. It did especially well in E/I and S/N classification.  Figure 5 showed the confusion matrix of the whole 16-type classification. The result for each type was normalized by the frequency of the type presented in our dataset to achieve better visualization.

% \FloatBarrier

\section{Conclusion}
We introduce a new user embedding framework, Author2Vec, based on BERT, which was pre-trained on an novel unsupervised objective: ``Authorship'' classification. Due to the abundance of the authorship training data on social media, this method provides a good basis for effectively utilizing user generated data to produce good user embedding that incorporates users' intrinsic attributes.

After pre-training Author2Vec on 10k randomly selected users, we performed a preliminary analysis by evaluating it on a simple gender classification task. The clear clusters shown in visualization and the classification F1-score of 93.3\% suggested that the Author2Vec embedding successfully encodes features of social media users.

We carried out experiments on two personality related benchmarks to further evaluate our user embedding model: depression detection and MBTI personality classification. User data in both datasets were carefully selected and filtered from Reddit forum. In both benchmarks, pre-trained Author2Vec embedding outperformed all the baseline embedding methods including LSI, LDA, and Word2Vec. This demonstrates that pre-trained Author2Vec is able to capture non-trivial intrinsic user features that can not be captured by traditional count-based and prediction-based methods.

\section{Future Work}
More work is required to utilise the proposed user embedding system. The content on social media platforms is noisy. This makes the data collection, data labelling and data analysis more difficult and tedious. Due to the difficulty in building reliable datasets, we currently only evaluated our embedding system on two benchmarks. However, more benchmarks are needed to further evaluate Author2Vec's ability to encode users' intrinsic properties.

Furthermore, the interpretation of the embedding representation remains shrouded. More work is required to investigate the underlying meaning of the user embedding.

We expect our Author2Vec to work well in social-media user related tasks without fine-tuning the entire embedding model on unseen documents. However, experiments using a large corpus might be required as in this paper we pre-trained our model using data only from approximately 10,000 authors on Reddit. Data from other social media platforms could also be tested in our framework to examine the generalizability of our Author2Vec embedding framework.

\bibliographystyle{acl_natbib}
\bibliography{mybibliography}

\begin{thebibliography}{43}
\expandafter\ifx\csname natexlab\endcsname\relax\def\natexlab#1{#1}\fi

\bibitem[{Abel et~al.(2011)Abel, Gao, Houben, and Tao}]{abel2011analyzing}
Fabian Abel, Qi~Gao, Geert-Jan Houben, and Ke~Tao. 2011.
\newblock Analyzing user modeling on twitter for personalized news
  recommendations.
\newblock In \emph{International Conference on User Modeling, Adaptation, and
  Personalization}, pages 1--12. Springer.

\bibitem[{Alowibdi et~al.(2013)Alowibdi, Buy, and Yu}]{alowibdi2013language}
Jalal~S Alowibdi, Ugo~A Buy, and Philip Yu. 2013.
\newblock Language independent gender classification on twitter.
\newblock In \emph{Proceedings of the 2013 IEEE/ACM international conference on
  advances in social networks analysis and mining}, pages 739--743. ACM.

\bibitem[{Benton et~al.(2016)Benton, Arora, and Dredze}]{benton2016learning}
Adrian Benton, Raman Arora, and Mark Dredze. 2016.
\newblock Learning multiview embeddings of twitter users.
\newblock In \emph{Proceedings of the 54th Annual Meeting of the Association
  for Computational Linguistics (Volume 2: Short Papers)}, pages 14--19.

\bibitem[{Bespalov et~al.(2011)Bespalov, Bai, Qi, and
  Shokoufandeh}]{bespalov2011sentiment}
Dmitriy Bespalov, Bing Bai, Yanjun Qi, and Ali Shokoufandeh. 2011.
\newblock Sentiment classification based on supervised latent n-gram analysis.
\newblock In \emph{Proceedings of the 20th ACM international conference on
  Information and knowledge management}, pages 375--382. ACM.

\bibitem[{Blei et~al.(2003)Blei, Ng, and Jordan}]{blei2003latent}
David~M Blei, Andrew~Y Ng, and Michael~I Jordan. 2003.
\newblock Latent dirichlet allocation.
\newblock \emph{Journal of machine Learning research}, 3(Jan):993--1022.

\bibitem[{Chen et~al.(2010)Chen, Nairn, Nelson, Bernstein, and
  Chi}]{chen2010short}
Jilin Chen, Rowan Nairn, Les Nelson, Michael Bernstein, and Ed~Chi. 2010.
\newblock Short and tweet: experiments on recommending content from information
  streams.
\newblock In \emph{Proceedings of the SIGCHI conference on human factors in
  computing systems}, pages 1185--1194. ACM.

\bibitem[{Cho et~al.(2014)Cho, Van~Merri{\"e}nboer, Bahdanau, and
  Bengio}]{cho2014properties}
Kyunghyun Cho, Bart Van~Merri{\"e}nboer, Dzmitry Bahdanau, and Yoshua Bengio.
  2014.
\newblock On the properties of neural machine translation: Encoder-decoder
  approaches.
\newblock \emph{arXiv preprint arXiv:1409.1259}.

\bibitem[{Chung et~al.(2014)Chung, Gulcehre, Cho, and
  Bengio}]{chung2014empirical}
Junyoung Chung, Caglar Gulcehre, KyungHyun Cho, and Yoshua Bengio. 2014.
\newblock Empirical evaluation of gated recurrent neural networks on sequence
  modeling.
\newblock \emph{arXiv preprint arXiv:1412.3555}.

\bibitem[{Deerwester et~al.(1990)Deerwester, Dumais, Furnas, Landauer, and
  Harshman}]{deerwester1990indexing}
Scott Deerwester, Susan~T Dumais, George~W Furnas, Thomas~K Landauer, and
  Richard Harshman. 1990.
\newblock Indexing by latent semantic analysis.
\newblock \emph{Journal of the American society for information science},
  41(6):391--407.

\bibitem[{Devlin et~al.(2018)Devlin, Chang, Lee, and
  Toutanova}]{devlin2018bert}
Jacob Devlin, Ming-Wei Chang, Kenton Lee, and Kristina Toutanova. 2018.
\newblock Bert: Pre-training of deep bidirectional transformers for language
  understanding.
\newblock \emph{arXiv preprint arXiv:1810.04805}.

\bibitem[{Ding et~al.(2017)Ding, Bickel, and Pan}]{ding2017multi}
Tao Ding, Warren~K Bickel, and Shimei Pan. 2017.
\newblock Multi-view unsupervised user feature embedding for social media-based
  substance use prediction.
\newblock In \emph{Proceedings of the 2017 Conference on Empirical Methods in
  Natural Language Processing}, pages 2275--2284.

\bibitem[{Gjurkovi{\'c} and {\v{S}}najder(2018)}]{gjurkovic2018reddit}
Matej Gjurkovi{\'c} and Jan {\v{S}}najder. 2018.
\newblock Reddit: A gold mine for personality prediction.
\newblock In \emph{Proceedings of the Second Workshop on Computational Modeling
  of People’s Opinions, Personality, and Emotions in Social Media}, pages
  87--97.

\bibitem[{Hochreiter and Schmidhuber(1997)}]{hochreiter1997long}
Sepp Hochreiter and J{\"u}rgen Schmidhuber. 1997.
\newblock Long short-term memory.
\newblock \emph{Neural computation}, 9(8):1735--1780.

\bibitem[{Hong and Davison(2010)}]{hong2010empirical}
Liangjie Hong and Brian~D Davison. 2010.
\newblock Empirical study of topic modeling in twitter.
\newblock In \emph{Proceedings of the first workshop on social media
  analytics}, pages 80--88. acm.

\bibitem[{Hu et~al.(2016)Hu, Xiao, Luo, and Nguyen}]{hu2016language}
Tianran Hu, Haoyuan Xiao, Jiebo Luo, and Thuy-vy~Thi Nguyen. 2016.
\newblock What the language you tweet says about your occupation.
\newblock In \emph{Tenth International AAAI Conference on Web and Social
  Media}.

\bibitem[{Kosinski et~al.(2013)Kosinski, Stillwell, and
  Graepel}]{kosinski2013private}
Michal Kosinski, David Stillwell, and Thore Graepel. 2013.
\newblock Private traits and attributes are predictable from digital records of
  human behavior.
\newblock \emph{Proceedings of the National Academy of Sciences},
  110(15):5802--5805.

\bibitem[{Le and Mikolov(2014)}]{le2014distributed}
Quoc Le and Tomas Mikolov. 2014.
\newblock Distributed representations of sentences and documents.
\newblock In \emph{International conference on machine learning}, pages
  1188--1196.

\bibitem[{Li et~al.(2019)Li, Yang, Xu, Wang, and Lin}]{li2019improving}
Yumeng Li, Liang Yang, Bo~Xu, Jian Wang, and Hongfei Lin. 2019.
\newblock Improving user attribute classification with text and social network
  attention.
\newblock \emph{Cognitive Computation}, pages 1--10.

\bibitem[{Maaten and Hinton(2008)}]{maaten2008visualizing}
Laurens van~der Maaten and Geoffrey Hinton. 2008.
\newblock Visualizing data using t-sne.
\newblock \emph{Journal of machine learning research}, 9(Nov):2579--2605.

\bibitem[{Makhzani and Frey(2013)}]{makhzani2013k}
Alireza Makhzani and Brendan Frey. 2013.
\newblock K-sparse autoencoders.
\newblock \emph{arXiv preprint arXiv:1312.5663}.

\bibitem[{Michelson and Macskassy(2010)}]{michelson2010discovering}
Matthew Michelson and Sofus~A Macskassy. 2010.
\newblock Discovering users' topics of interest on twitter: a first look.
\newblock In \emph{Proceedings of the fourth workshop on Analytics for noisy
  unstructured text data}, pages 73--80. ACM.

\bibitem[{Mikolov et~al.(2013)Mikolov, Chen, Corrado, and
  Dean}]{mikolov2013efficient}
Tomas Mikolov, Kai Chen, Greg Corrado, and Jeffrey Dean. 2013.
\newblock Efficient estimation of word representations in vector space.
\newblock \emph{arXiv preprint arXiv:1301.3781}.

\bibitem[{Mitchell et~al.(2015)Mitchell, Hollingshead, and
  Coppersmith}]{mitchell2015quantifying}
Margaret Mitchell, Kristy Hollingshead, and Glen Coppersmith. 2015.
\newblock Quantifying the language of schizophrenia in social media.
\newblock In \emph{Proceedings of the 2nd workshop on Computational linguistics
  and clinical psychology: From linguistic signal to clinical reality}, pages
  11--20.

\bibitem[{Mueller and Stumme(2016)}]{mueller2016gender}
Juergen Mueller and Gerd Stumme. 2016.
\newblock Gender inference using statistical name characteristics in twitter.
\newblock In \emph{Proceedings of the The 3rd Multidisciplinary International
  Social Networks Conference on SocialInformatics 2016, Data Science 2016},
  page~47. ACM.

\bibitem[{Myers et~al.(1998)Myers, McCaulley, Quenk, and
  Hammer}]{myers1998mbti}
Isabel~Briggs Myers, Mary~H McCaulley, Naomi~L Quenk, and Allen~L Hammer. 1998.
\newblock \emph{MBTI manual: A guide to the development and use of the
  Myers-Briggs Type Indicator}, volume~3.
\newblock Consulting Psychologists Press Palo Alto, CA.

\bibitem[{Ofoghi et~al.(2016)Ofoghi, Mann, and Verspoor}]{ofoghi2016towards}
Bahadorreza Ofoghi, Meghan Mann, and Karin Verspoor. 2016.
\newblock Towards early discovery of salient health threats: A social media
  emotion classification technique.
\newblock In \emph{Biocomputing 2016: Proceedings of the Pacific Symposium},
  pages 504--515. World Scientific.

\bibitem[{Pennacchiotti and Popescu(2011)}]{pennacchiotti2011machine}
Marco Pennacchiotti and Ana-Maria Popescu. 2011.
\newblock A machine learning approach to twitter user classification.
\newblock In \emph{Fifth International AAAI Conference on Weblogs and Social
  Media}.

\bibitem[{Pennebaker et~al.(2015)Pennebaker, Boyd, Jordan, and
  Blackburn}]{pennebaker2015development}
James~W Pennebaker, Ryan~L Boyd, Kayla Jordan, and Kate Blackburn. 2015.
\newblock The development and psychometric properties of liwc2015.
\newblock Technical report.

\bibitem[{Pennington et~al.(2014)Pennington, Socher, and
  Manning}]{pennington2014glove}
Jeffrey Pennington, Richard Socher, and Christopher Manning. 2014.
\newblock Glove: Global vectors for word representation.
\newblock In \emph{Proceedings of the 2014 conference on empirical methods in
  natural language processing (EMNLP)}, pages 1532--1543.

\bibitem[{Pirina and {\c{C}}{\"o}ltekin(2018)}]{pirina2018identifying}
Inna Pirina and {\c{C}}a{\u{g}}r{\i} {\c{C}}{\"o}ltekin. 2018.
\newblock Identifying depression on reddit: The effect of training data.
\newblock In \emph{Proceedings of the 2018 EMNLP Workshop SMM4H: The 3rd Social
  Media Mining for Health Applications Workshop \& Shared Task}, pages 9--12.

\bibitem[{Preo{\c{t}}iuc-Pietro et~al.(2015)Preo{\c{t}}iuc-Pietro, Eichstaedt,
  Park, Sap, Smith, Tobolsky, Schwartz, and Ungar}]{preoctiuc2015role}
Daniel Preo{\c{t}}iuc-Pietro, Johannes Eichstaedt, Gregory Park, Maarten Sap,
  Laura Smith, Victoria Tobolsky, H~Andrew Schwartz, and Lyle Ungar. 2015.
\newblock The role of personality, age, and gender in tweeting about mental
  illness.
\newblock In \emph{Proceedings of the 2nd workshop on computational linguistics
  and clinical psychology: From linguistic signal to clinical reality}, pages
  21--30.

\bibitem[{Rahimi et~al.(2015)Rahimi, Vu, Cohn, and
  Baldwin}]{rahimi2015exploiting}
Afshin Rahimi, Duy Vu, Trevor Cohn, and Timothy Baldwin. 2015.
\newblock Exploiting text and network context for geolocation of social media
  users.
\newblock \emph{arXiv preprint arXiv:1506.04803}.

\bibitem[{Salton and Buckley(1988)}]{salton1988term}
Gerard Salton and Christopher Buckley. 1988.
\newblock Term-weighting approaches in automatic text retrieval.
\newblock \emph{Information processing \& management}, 24(5):513--523.

\bibitem[{Schwartz et~al.(2013)Schwartz, Eichstaedt, Kern, Dziurzynski,
  Ramones, Agrawal, Shah, Kosinski, Stillwell, Seligman
  et~al.}]{schwartz2013personality}
H~Andrew Schwartz, Johannes~C Eichstaedt, Margaret~L Kern, Lukasz Dziurzynski,
  Stephanie~M Ramones, Megha Agrawal, Achal Shah, Michal Kosinski, David
  Stillwell, Martin~EP Seligman, et~al. 2013.
\newblock Personality, gender, and age in the language of social media: The
  open-vocabulary approach.
\newblock \emph{PloS one}, 8(9):e73791.

\bibitem[{Sesa-Nogueras et~al.(2016)Sesa-Nogueras, Faundez-Zanuy, and
  Roure-Alcob{\'e}}]{sesa2016gender}
Enric Sesa-Nogueras, Marcos Faundez-Zanuy, and Josep Roure-Alcob{\'e}. 2016.
\newblock Gender classification by means of online uppercase handwriting: a
  text-dependent allographic approach.
\newblock \emph{Cognitive Computation}, 8(1):15--29.

\bibitem[{Sloan et~al.(2015)Sloan, Morgan, Burnap, and
  Williams}]{sloan2015tweets}
Luke Sloan, Jeffrey Morgan, Pete Burnap, and Matthew Williams. 2015.
\newblock Who tweets? deriving the demographic characteristics of age,
  occupation and social class from twitter user meta-data.
\newblock \emph{PloS one}, 10(3):e0115545.

\bibitem[{Song and Lee(2017)}]{song2017learning}
Yan Song and Chia-Jung Lee. 2017.
\newblock Learning user embeddings from emails.
\newblock In \emph{Proceedings of the 15th Conference of the European Chapter
  of the Association for Computational Linguistics: Volume 2, Short Papers},
  volume~2, pages 733--738.

\bibitem[{Volkova et~al.(2015)Volkova, Bachrach, Armstrong, and
  Sharma}]{volkova2015inferring}
Svitlana Volkova, Yoram Bachrach, Michael Armstrong, and Vijay Sharma. 2015.
\newblock Inferring latent user properties from texts published in social
  media.
\newblock In \emph{Twenty-Ninth AAAI Conference on Artificial Intelligence}.

\bibitem[{Xiao(2018)}]{xiao2018bertservice}
Han Xiao. 2018.
\newblock bert-as-service.
\newblock \url{https://github.com/hanxiao/bert-as-service}.

\bibitem[{Xu et~al.(2012)Xu, Zhang, Wu, and Yang}]{xu2012modeling}
Zhiheng Xu, Yang Zhang, Yao Wu, and Qing Yang. 2012.
\newblock Modeling user posting behavior on social media.
\newblock In \emph{Proceedings of the 35th international ACM SIGIR conference
  on Research and development in information retrieval}, pages 545--554. ACM.

\bibitem[{Yang et~al.(2019)Yang, Dai, Yang, Carbonell, Salakhutdinov, and
  Le}]{yang2019xlnet}
Zhilin Yang, Zihang Dai, Yiming Yang, Jaime Carbonell, Ruslan Salakhutdinov,
  and Quoc~V Le. 2019.
\newblock Xlnet: Generalized autoregressive pretraining for language
  understanding.
\newblock \emph{arXiv preprint arXiv:1906.08237}.

\bibitem[{Yu et~al.(2016)Yu, Wan, and Zhou}]{yu2016user}
Yang Yu, Xiaojun Wan, and Xinjie Zhou. 2016.
\newblock User embedding for scholarly microblog recommendation.
\newblock In \emph{Proceedings of the 54th Annual Meeting of the Association
  for Computational Linguistics (Volume 2: Short Papers)}, pages 449--453.

\bibitem[{Zhang et~al.(2018)Zhang, Wang, Wang, and Zha}]{zhang2018user}
Wei Zhang, Wen Wang, Jun Wang, and Hongyuan Zha. 2018.
\newblock User-guided hierarchical attention network for multi-modal social
  image popularity prediction.
\newblock In \emph{Proceedings of the 2018 World Wide Web Conference}, pages
  1277--1286. International World Wide Web Conferences Steering Committee.

\end{thebibliography}

\appendix

\end{document}